\documentclass[11pt,a4paper]{article}
\usepackage[hyperref]{ranlp2021}
\usepackage{times}
\usepackage{latexsym}

\usepackage{graphicx} % DO NOT CHANGE THIS
\usepackage{amsmath}
\usepackage{nccmath}
\usepackage{amssymb}
\usepackage{mathtools}
\usepackage{times}
\usepackage{xcolor}
\usepackage{multirow}

\usepackage{microtype}

\aclfinalcopy % Uncomment this line for the final submission
\title{A Hierarchical Entity Graph Convolutional Network \\for Relation Extraction across Documents}

\author{Tapas Nayak\thanks{$\quad$This work was done when the first author was a PhD student at the National University of Singapore.} \\
  Department of Computer Science \\
  Indian Institute of Technology Kharagpur \\
  \texttt{tnk02.05@gmail.com} \\\And
  Hwee Tou Ng \\
  Department of Computer Science \\
  National University of Singapore \\
  \texttt{nght@comp.nus.edu.sg} \\}
  
\date{}

\begin{document}
\maketitle

\begin{abstract}

Distantly supervised datasets for relation extraction mostly focus on sentence-level extraction, and they cover very few relations. In this work, we propose cross-document relation extraction, where the two entities of a relation tuple appear in two different documents that are connected via a chain of common entities. Following this idea, we create a dataset for two-hop relation extraction, where each chain contains exactly two documents. Our proposed dataset covers a higher number of relations than the publicly available sentence-level datasets. We also propose a hierarchical entity graph convolutional network (HEGCN) model for this task that improves performance by 1.1\% F1 score on our two-hop relation extraction dataset, compared to some strong neural baselines. 

\end{abstract}

\section{Introduction}

The idea of distant supervision \citep{mintz2009distant} eliminates the need for manual annotation for obtaining training data for relation extraction. Previously, this idea is used mostly to create sentence-level datasets. However, the assumption of distant supervision, that the two entities of a tuple must appear in the same sentence, is overly strict. We may not find an adequate number of evidence sentences for many relations as both entities do not appear in the same sentence. The relation extraction models built on such data can find relations only for a small number of relations and the relations of most knowledge bases (KBs) will be out of the reach of such models.

To address this issue, we propose a multi-hop relation extraction task where the subject and object entities of a tuple can appear in two different documents, and these two documents are connected via some common entities. We can create a chain of entities from the subject entity to the object entity of a tuple via the common entities across multiple documents. Each link in this chain represents a relation between the entities located at the endpoints of the link. We can determine the relation between the subject and object entities of a tuple by following this chain of relations. This approach can give training instances for more relations than sentence-level distant supervision. Following the proposed multi-hop approach, we create a two-hop relation extraction dataset for the task. Each instance of this dataset has two documents, where the first document contains the subject entity and the second document contains the object entity of a tuple. These two documents are connected via at least one common entity. This idea can be extended to create an N-hop dataset.

We also propose a hierarchical entity graph convolutional network (HEGCN) model for the task. Our proposed model has two levels of graph convolutional networks (GCNs). The first-level GCN of the hierarchy is applied to the entity mention level graph of every document to capture the relations among the entity mentions within a document. The second-level GCN of the hierarchy is applied on a unified entity-level graph, which is built using all the unique entities present in the document chain. This entity-level graph can be built on the document chain of any length and it can capture the relations among the entities across the multiple documents in the chain. Our proposed HEGCN model improves the performance on our two-hop dataset. To summarize, the following are the contributions of this paper:

(1) We propose a multi-hop relation extraction task and create a two-hop dataset. This dataset has more relations than other popular distantly supervised sentence-level or document-level relation extraction datasets. 

(2) We propose a novel hierarchical entity graph convolutional network (HEGCN) for multi-hop relation extraction. Our proposed model improves the F1 score by 1.1\% on our two-hop dataset, compared to strong neural baselines\footnote{The source code and data for this paper are available at https://github.com/nusnlp/MHRE.git}.

\section{Task Formalization}

Multi-hop relation extraction can be defined as follows. Consider two entities, a subject entity $e_s$ and an object entity $e_o$, and a chain of documents $D=\{D_s \rightarrow D_1 \rightarrow D_2 \rightarrow ... \rightarrow D_n \rightarrow D_o\}$ where $e_s \in D_s$ and $e_o \in D_o$. There exists a chain of entities $e_s \rightarrow c_1 \rightarrow c_2 \rightarrow ... \rightarrow c_{n+1} \rightarrow e_o$ where $c_1 \in \{D_s, D_1\}$, $c_2 \in \{D_1, D_2\}, ... , $ $c_{n+1} \in \{D_{n}, D_o\}$. The task is to find the relation between $e_s$ and $e_o$ from a pre-defined set of relations $R \cup \{\mathit{None}\}$, where $R$ is the set of relations and {\em None} indicates that none of the relations in $R$ holds between $e_s$ and $e_o$. A simpler version of this task is two-hop relation extraction where $D_s$ and $D_o$ are directly connected by at least one common entity. In this paper, we focus on two-hop relation extraction.

\section{Related Work}

\subsection{Relation Extraction Datasets}

Distantly supervised datasets are very popular for relation extraction \cite{Nayak2021DeepNA}. \citet{riedel2010modeling} (NYT10) and \citet{hoffmann2011knowledge} (NYT11) mapped Freebase tuples to New York Times (NYT) articles to obtain such datasets. The NYT10 and NYT11 datasets have been used extensively by researchers for relation extraction. TACRED \cite{zhang2017position} is another dataset created from the TAC KBP evaluations. FewRel 2.0 \cite{gao2019fewrel} is a few-shot relation extraction dataset. All these datasets are created at the sentence level. DocRED \cite{yao2019DocRED} is a document-level relation extraction dataset created using Wikipedia articles and Wikidata items. To the best of our knowledge, there does not exist any relation extraction dataset which involves multiple documents.

\subsection{Relation Extraction Models}

Neural models have performed well on distantly supervised datasets for relation extraction. \citet{zeng2014relation,zeng2015distant} used convolutional network with max-pooling on word embeddings for this task, whereas \citet{huang2016attention,jat2018attention,nayak2019effective} used word-level attention model for single-instance sentence-level relation extraction. \citet{lin2016neural,vashishth2018reside,ye2019distant} used neural networks in a multi-instance setting to find a relation from a bag of independent sentences. Recently, graph convolutional network-based (GCN) \cite{Kipf2017SemiSupervisedCW} models have become popular for many NLP tasks. These models work on non-linear graph structures. \citet{zhang2018graph,vashishth2018reside,guo2019aggcn,Zeng2020DoubleGB} used graph convolution networks for relation extraction. They consider each token in a sentence as a node in the graph and use a syntactic dependency tree to create a graph structure among the nodes. Recently, neural joint extraction approaches \cite{takanobu2019hrlre,nayak2019ptrnetdecoding} were proposed for this task.

\subsection{Multi-hop QA versus Multi-hop RE}

\citet{welbl2018constructing} proposed a multi-hop QA dataset (WikiHop) where the answer can only be found using more than one document. Several neural models have been proposed \cite{song2018exploring,cao2019bag,de2019question,kundu2019exploiting} to solve this task. We have created a two-hop relation extraction dataset (THRED) from this WikiHop dataset. The major difference between these two datasets is that THRED contains many {\em None} relations, whereas in the WikiHop dataset, every instance has a correct answer. Extracting the {\em None} relation is challenging, since {\em None} occurs when no relations in $R$ exist. When the number of relations in $R$ increases, it becomes more difficult to predict the relations. As such, we believe the multi-hop RE task is more challenging than the multi-hop QA task.

\section{Dataset Construction}

We create a two-hop relation extraction dataset from a multi-hop question-answering (QA) dataset WikiHop \cite{welbl2018constructing}. \citet{welbl2018constructing} defined the multi-hop QA task as follows: Given a set of supporting documents $D_s$ and a set of candidate answers $C_a$ which are mentioned in $D_s$, the goal is to find the correct answer $a^* \in C_a$ for a question by drawing on the supporting documents. They used Wikipedia articles and Wikidata \cite{vrandevcic2014wikidata} tuples for creating this dataset. Each positive tuple $(e_s, e_o, r_p)$ in Wikidata has two entities, a subject entity $e_s$ and an object entity $e_o$, and a positive relation $r_p$ between the subject and object entity. The questions are created by combining the subject entity $e_s$ and the relation $r_p$, and the object entity $e_o$ is the correct answer $a^*$ for a given question. The other candidate answers are carefully chosen from Wikidata entities so that they have a similar type as the correct answer. The supporting documents are chosen in such a way that at least two documents are needed to find the correct answer. This means the subject entity $e_s$ and the object entity $e_o$ do not appear in the same document. They used a bipartite graph partition technique to create the dataset. In this bipartite graph, vertices on one side correspond to Wikidata entities, and vertices on the other side correspond to Wikipedia articles. An edge is created between an entity vertex and a document vertex if this document contains the entity. As we traverse the graph starting from vertex $e_s$, it visits many document vertices and entity vertices. This constitutes the supporting document set and candidate answer set. If the candidate answer set does not contain the object entity $e_o$ which is the correct answer, this instance is discarded. They also limited the length of the traversal to three documents. \citet{welbl2018constructing} only released the supporting documents, questions, and candidate answers for their dataset. They did not release the connecting entities.

We convert this WikiHop dataset into a two-hop relation extraction dataset. The subject entities and the candidate entities can be easily found in the documents using string matching. We use a named entity recognizer from spaCy\footnote{https://spacy.io/} to find the other entities in the documents and these entities can link these documents. We find that most of the WikiHop question-answer instances are two-hop instances. That means for most of the instances of WikiHop dataset, there is at least one document pair in the supporting document set where the first document of the pair contains the subject entity and the second document of the pair contains the correct answer, and these two documents in the pair are directly connected via some third entity. To simplify the multi-hop relation extraction task, we fix the hop count at 2. For every instance of the WikiHop dataset, we can easily find the subject entity $e_s$ and the positive relation $r_p$ from the question. The correct answer $a^*$ is the object entity of a positive tuple. $(e_s, a^*, r_p)$ is the positive tuple for relation extraction. For any other candidate answer $e_w \in C_a-\{a^*\}$, the entity pair $(e_s, e_w)$ is considered as a {\em None} tuple if there exists no relation among the four pairs $(e_s, e_w)$, $(e_w, e_s)$, $(e_w, e_o)$, and $(e_o, e_w)$ in Wikidata. We check for the no relation condition for these four entity pairs involving $e_w$, $e_s$, and $e_o$ to reduce the distant supervision noise in the dataset for {\em None} tuples. We create a {\em None} candidate set $C_n$ with each $e_w \in C_a-\{a^*\}$. We first find all possible pairs of documents from the supporting document set $D_s$ such that the first document of the pair contains the subject entity $e_s$ and the second document of the pair contains either the entity $a^*$ or one of the entities from $C_n$. We discard those pairs of documents that do not contain any common entity. The document pairs where the second document contains the entity $a^*$ are considered as a document chain for the positive tuple $(e_s, a^*, r_p)$ where $r_p \in R$. All other document pairs where the second document contains an entity from the set $C_n$ are considered as a document chain for {\em None} tuple $(e_s, e_w, None)$ where $e_w \in C_n$. In this way, using distant supervision, we can create a dataset for two-hop relation extraction. Each instance of this dataset has a chain of documents $D=\{D_s \rightarrow D_o\}$ of length 2 that is the textual source of a tuple $(e_s, e_o, r)$. The document $D_s$ contains the subject entity $e_s$ and the document $D_o$ contains the object entity $e_o$. The two documents are connected with at least one common entity $c$. There exists at least one entity chain $e_s \rightarrow c \rightarrow e_o$ in the document chain. The goal is to find the relation $r$ between $e_s$ and $e_o$ from the set $R \cup \{\mathit{None}\}$. We refer to this two-hop dataset as THRED (two-hop relation extraction dataset) in the remaining sections of this paper. We manually checked 100 randomly selected positive samples and 100 randomly selected negative samples, and found that 76\% of the selected positive samples and 82\% of the selected negative samples are accurate.

% Negative: 82\%
% Positive: 76\%

\begin{table}[ht]
\small
\centering
\begin{tabular}{|l|l|}
\hline
Question      & located\_in\_administrative\_entity Zoo Lake                                                                                                                                                                                                                                                                                                                                                                                                                                                                                                                  \\ \hline
Candidates & Gauteng, Tanzania                                                                                                                                                                                                                                                                                                                                                                                                                                                                               \\ \hline
Answer     & Gauteng                                                                                                                                                                                                                                                                                                                                                                                                                                                                                                                           \\ \hline
Doc1       & \begin{tabular}[c]{@{}l@{}}\textcolor{red}{\textbf{Zoo Lake}} is a popular lake and public \\park in \textcolor{orange}{\textbf{Johannesburg}} , \textcolor{orange}{\textbf{South Africa}} . \\It is part of the \textbf{Hermann Eckstein Park} \\and is opposite the \textbf{Johannesburg Zoo} . \\The \textcolor{red}{\textbf{Zoo Lake}} consists of two dams , \\an upper feeder dam , and a larger lower \\dam , both constructed in natural \\marshland watered by the\\ \textbf{Parktown Spruit} .\end{tabular}                                                                                                                                                       \\ \hline
Doc2       & \begin{tabular}[c]{@{}l@{}}\textcolor{orange}{\textbf{Johannesburg}} is the largest city in \\\textcolor{orange}{\textbf{South Africa}} and is one of the 50 \\largest urban areas in the world . It \\is the provincial capital of \textcolor{blue}{\textbf{Gauteng}} , \\which is the wealthiest province in\\ \textcolor{orange}{\textbf{South Africa}} . \end{tabular} \\\hline

Doc3       & \begin{tabular}[c]{@{}l@{}}\textbf{Mozambique} is a country in \textbf{Southeast} \\\textbf{Africa} bordered by the \textbf{Indian Ocean} to \\the east , \textcolor{blue}{\textbf{Tanzania}} to the north , \textbf{Malawi} \\and \textbf{Zambia} to the northwest , \textbf{Zimbabwe} \\to the west , and \textbf{Swaziland} and \textcolor{orange}{\textbf{South}}\\ \textcolor{orange}{\textbf{Africa}} to the southwest . \end{tabular}                                                                                                                                                                                                                                                                                                         \\ \hline
\end{tabular}
\caption{A multi-hop question-answer instance from the WikiHop dataset. The tuple (Doc1, {\em Zoo Lake}, Doc2, {\em Gauteng}, {\em located\_in\_administrative\_entity}) constitutes a positive instance in the THRED dataset. The tuple (Doc1, {\em Zoo Lake}, Doc3, {\em Tanzania}, {\em None}) constitutes a negative instance in the THRED dataset.}
\label{tab:wikihop}
\end{table}

\begin{table}[ht]
\small
\centering
\begin{tabular}{l|ccc}
\hline
                     & \multicolumn{1}{c}{Train} & \multicolumn{1}{c}{Test} \\ \hline
\#Positive relations    & 218       & 72       \\ 
\#Document chains    & 143,906     & 5,320          \\ 
\#Positive instances & 40,247      & 1,672          \\
\#Positive entity pairs & 21,490   & 618             \\
\#None instances     & 197,731     & 7,806          \\ \hline
\end{tabular}
\caption{Statistics of the THRED dataset.}
\label{tab:dataset_stat}
\end{table}

\subsection{Dataset Statistics}

The training, validation, and test data of the WikiHop dataset are created using distant supervision, but the validation and test data are manually verified. WikiHop test data is blind and not released. So we use their validation data to create the test data for our task and use their training data for our training and validation purposes. We include the statistics of our two-hop relation extraction dataset in Table \ref{tab:dataset_stat}. We include the statistics on the number of common entities present in the two documents of a chain in Table \ref{tab:common_entity_stat}. We split the training data randomly, with 90\% for training and 10\% for validation. From Table \ref{tab:dataset_stat}, we see that the dataset contains a much higher number of {\em None} tuples than the positive tuples. So we randomly select {\em None} tuples so that the number of {\em None} tuples is the same as the number of positive tuples for training and validation. For evaluation, we consider the entire test dataset. From Table \ref{tab:relations}, we see that our THRED dataset contains more relations than any other distantly supervised relation extraction datasets such as the New York Times \cite{riedel2010modeling,hoffmann2011knowledge} or DocRED \cite{yao2019DocRED}.

\begin{table}[ht]
\small
\centering
\begin{tabular}{c|cc}
\hline
\multicolumn{1}{l|}{} & \multicolumn{2}{l}{\#Document chains} \\ \hline
\#Common entities      & Train              & Test              \\ \hline
1                      & 92,140             & 3,615             \\ 
2                      & 36,275             & 1,161             \\ 
3                      & 10,824             & 374               \\ 
4                      & 3,170              & 113               \\ 
$\geq$5                 & 1,497              & 57                \\ \hline
\end{tabular}
\caption{Statistics of the common entities in the THRED dataset.}
\label{tab:common_entity_stat}
\end{table}

\begin{table}[ht]
\small
\centering
\begin{tabular}{l|c|l|c}
\hline
Dataset            & $\vert R \vert$ & Dataset & $\vert R \vert$ \\ \hline
NYT10              & 53     & NYT11              & 24   \\ 
TACRED             & 41   & DocRED             & 96     \\ 
FewRel 2.0  & 100 & \textbf{THRED}              & \textbf{218}   \\ \hline
\end{tabular}
\caption{The number of relations in various relation extraction datasets. $R$ is the set of positive relations.}
\label{tab:relations}
\end{table}

\section{Proposed HEGCN Model}

\begin{figure*}[ht]
\centering
\includegraphics[scale=0.45]{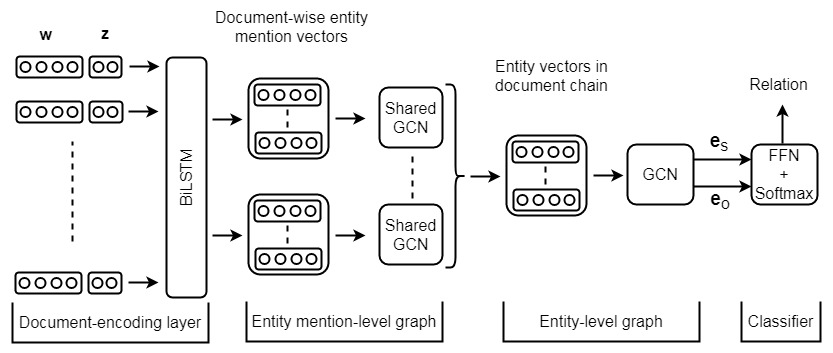}
\caption{The architecture of our proposed HEGCN model. GCN in entity mention-level graph is shared across the documents in a chain. This diagram is for document chain of length 2.}
\label{fig:hegcn}
\end{figure*}

We propose a hierarchical entity graph convolutional network (HEGCN) for multi-hop relation extraction. We encode the documents in a document chain using a bi-directional long short-term memory (BiLSTM) layer \cite{hochreiter1997long}. On top of the BiLSTM layer, we use two graph convolutional networks (GCN), one after another in a hierarchy. In the first level of the GCN hierarchy, we construct a separate entity mention graph on each document of the chain using all the entities mentioned in that document. Each mention of an entity in a document is considered as a separate node in the graph. We use a graph convolutional network (GCN) to represent the entity mention graph of each document to capture the relations among the entity mentions in the document. We then construct a unified entity-level graph across all the documents in the chain. Each node of this entity-level graph represents a unique entity in the document chain. Each common entity between two documents in the chain is represented by a single node in the graph. We use a GCN to represent this entity-level graph to capture the relations among the entities across the documents. We concatenate the representations of the nodes of the subject entity and object entity and pass it to a feed-forward layer with softmax for relation classification.

\subsection{Documents Encoding Layer}

We use two types of embedding vectors: (1) word embedding vector $\mathbf{w} \in \mathbb{R}^{d_w}$ (2) entity token indicator embedding vector $\mathbf{z} \in \mathbb{R}^{d_z}$, which indicates if a word belongs to the subject entity, object entity, or common entities. The subject and object entities are assigned the embedding index of $2$ and $3$, respectively. The common entities in the document chain are assigned embedding index in an increasing order starting from index $4$. The same entities present in two documents in the chain get the same embedding index. Embedding index $0$ is used for padding and $1$ is used for all other tokens in the documents. A document is represented using a sequence of vectors  $\{\mathbf{x}_1, \mathbf{x}_2,....., \mathbf{x}_n\}$ where $\mathbf{x}_t = \mathbf{w}_t \Vert \mathbf{z}_t$. $\Vert$ represents the concatenation of vectors and $n$ is the document length. We concatenate all documents in a chain sequentially by using a document separator token. These token vectors are passed to a BiLSTM layer to capture the interaction among the documents in a chain. $\overrightarrow{\mathbf{h}_t} \in \mathbb{R}^{(d_w+d_z)}$ and $\overleftarrow{\mathbf{h}_t} \in \mathbb{R}^{(d_w+d_z)}$ are the output at the $t$th step of the forward LSTM and backward LSTM respectively. We concatenate them to obtain the $t$th BiLSTM output $\mathbf{h}_t \in \mathbb{R}^{2(d_w+d_z)}$.

\subsection{Hierarchical Entity Graph Convolutional Layers}

\citet{Kipf2017SemiSupervisedCW} proposed graph convolutional networks (GCN) which work on graph structures. Here, we describe the GCN which is used in our model. We represent a graph $\mathcal{G}$ with $m$ nodes using an adjacency matrix $\mathbf{A}$ of size $m \times m$. If there is an edge between node $i$ and node $j$, then $A_{ij} = A_{ji} = 1$. We also add self loops, $A_{ii} = 1$, in the graph $\mathcal{G}$. We normalize the adjacency matrix $\mathbf{A}$ by using symmetric normalization proposed by \citet{Kipf2017SemiSupervisedCW}. A diagonal node degree matrix $\mathbf{D}$ of size $m \times m$ is used in the normalization of $\mathbf{A}$. $\text{deg}(v_i)$ is the number of edges that are connected to the node $v_i$ in $\mathcal{G}$ and $\hat{\mathbf{A}}$ is the corresponding normalized adjacency matrix of $\mathcal{G}$. Each node of the graph receives the hidden representation of its neighboring nodes from the $(l-1)$th layer and uses the following operation to update its own hidden representation.

\begin{align*}
%     &D_{ij}=\begin{cases}
%     \text{deg}(v_i) & \text{if } i=j\\
%     0 & \text{otherwise}
% \end{cases}\\
%     &D^{-\frac{1}{2}}_{ij} = \begin{cases}
%     \frac{1}{\sqrt{D}_{ij}} & \text{if } i=j\\
%     0 & \text{otherwise}
% \end{cases}\\
&D^{-\frac{1}{2}}_{ij} = \begin{cases}
    \frac{1}{\sqrt{\text{deg}(v_i)}} & \text{if } i=j\\
    0 & \text{otherwise}
\end{cases}
\end{align*}
\begin{align*}
    &\hat{\mathbf{A}} = \mathbf{D}^{-\frac{1}{2}} \mathbf{A} \mathbf{D}^{-\frac{1}{2}}\\
    &\textbf{g}_i^l = \text{ReLU}(\sum_{j=1}^m \hat{A}_{ij} \textbf{W}^l \textbf{g}_j^{l-1})
\end{align*}

\noindent $\textbf{W}^l$ is the trainable weight matrix of the $l$th layer of the GCN, $\textbf{g}_i^l$ is the representation of the $i$th node of the graph at the $l$th layer. If $\textbf{g}_i^l$ has the dimension of $d_g$, then the dimension of the weight matrix $\textbf{W}^l$ is $d_g \times d_g$. $\textbf{g}_i^0$ is the initial input to the GCN.

\subsubsection{Entity Mention Graph Layer}

We construct an entity mention graph (EMG) for each document in the chain on top of the document encoding layer. An entity string may appear at multiple locations in a document and each appearance is considered as an entity mention. We add a node in the graph for each entity mention. We connect two entity mention nodes if they appear in the same sentence (EMG type 1 edge). We assume that since they appear in the same sentence, there may exist some relation between them. We also connect two entity mention nodes if the strings of the two entity mentions are identical (EMG type 2 edge). Let $e_1, \ldots, e_l$ be the sequence of entity mention nodes listed in the order of their appearance in a document. We connect nodes $e_i$ and $e_{i+1}$ ($1 \leq i < l$) with an edge (EMG type 3 edge). EMG type 3 edges create a linear chain of the entity mentions and ensure that the graph is connected. We use a graph convolutional network on this graph topology to capture the relations among the entity mentions in a document.

We obtain the initial representations of the entity mention nodes from the hidden representations of the document encoding layer. We concatenate the hidden vector of the first token of an entity mention, the hidden vector of its last token, and a context vector to obtain the entity mention node representation. The context vector is obtained using an attention mechanism on the tokens of the sentence in which the entity mention appears. 
\begin{align*}
    &\textbf{p} = \textbf{h}_b ~\Vert~ \textbf{h}_e, \quad
    s_t = \text{tanh}(\textbf{p}^T \textbf{W}) \textbf{h}_t\\
    &\textbf{a} = \text{softmax}({[s_1  s_2  \ldots  s_k]}^T)\\
    &\textbf{c} =  \sum_{t=1}^{k} \text{a}_t \textbf{h}_t, \quad
    \textbf{q} = \textbf{p} ~\Vert~ \textbf{c}
\end{align*}
\noindent $\textbf{h}_b \in \mathbb{R}^{2(d_w+d_z)}$ and $\textbf{h}_e \in \mathbb{R}^{2(d_w+d_z)}$ are the hidden vectors from the document encoding layer of the first and last token of an entity mention. $\textbf{W} \in \mathbb{R}^{4(d_w+d_z) \times 2(d_w+d_z)}$ is a trainable weight matrix, $\textbf{h}_t \in \mathbb{R}^{2(d_w+d_z)}$ is the hidden vector of the $t$th token of the sentence in which the entity mention is located, and $\text{a}_t$ is the normalized attention score for the $t$th token with respect to the entity mention. $k$ is the length of the sentence in which the entity mention is located, and $\textbf{c} \in \mathbb{R}^{2(d_w+d_z)}$ is the context vector. The entity mention node vector $\textbf{q} \in \mathbb{R}^{6(d_w+d_z)}$ of the $i$th node in the graph is passed to the GCN as $\textbf{g}_i^0$. The parameters of this GCN are shared across the documents in a chain. This layer of the model is referred to as entity mention-level graph convolutional network or EMGCN.

\subsubsection{Entity Graph Layer}

We construct a unified entity graph (EG) on top of the entity mention graphs. First, we construct an entity graph for each document, where each unique entity string is represented as an entity node in the graph. We add an edge between two entity nodes if the strings of the two entities appear together in at least one sentence in the document (EG type 1 edge). We also form a sequence of entity nodes based on the order of appearance of the entities in a document, where only the first occurrence of multiple occurrences of an entity is kept in the sequence. We connect two consecutive entity nodes in the sequence with an edge (EG type 2 edge). This ensures that the entire entity graph remains connected. 

We construct one entity graph for each document in the document chain. We unify the entity graphs of multiple documents by merging the nodes of common entities between them. The unified entity graph contains all the nodes from the multiple entity graphs, but the common entity nodes which appear in two entity graphs are merged into one node in the unified graph. There is an edge between two entity nodes in the unified entity graph if there exists an edge between them in any of the entity graphs of the documents.

We obtain the initial representations of the entity nodes from the GCN outputs of the entity mention graphs. For the common entities between two documents, we average the GCN outputs of the entity mention nodes that have an identical string as the entity from the entity mention graphs of the two documents. For other entity nodes that appear only in one document, we average the GCN outputs of the entity mention nodes that have an identical string as the entity from the entity mention graph of that document. Each entity vector is passed to another graph convolutional network as $\textbf{g}_i^0$ which represents the initial representation of the $i$th entity node in the unified entity graph. We use a graph convolutional network on this graph topology to capture the relations among the entities across the documents in the document chain. This layer of the model is referred to as entity-level graph convolutional network or EGCN. 

\begin{figure*}[ht]
\centering
\includegraphics[scale=0.28]{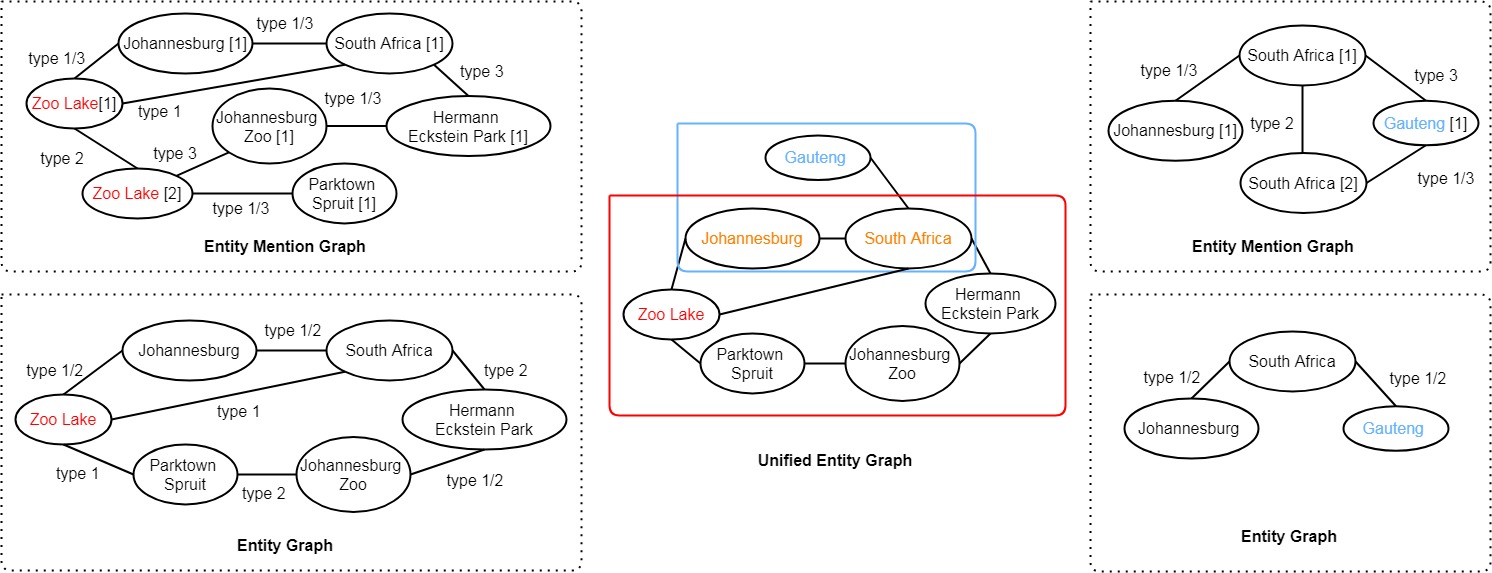}
\caption{The graph construction process for the positive instance in Table \ref{tab:wikihop}. The entity mention graph and entity graph on the left are for Doc1. The entity mention graph and entity graph on the right are for Doc2. The numbers in square brackets ([x]) in the entity mention graph are used to distinguish the entity mentions with identical string. Type x/y means this edge can be of both type x and type y. The `EMG' and `EG' prefixes are omitted from the labels of the edges in the entity mention graph and entity graph respectively. The unified entity graph is shown in the middle. Nodes in the red box are part of the entity graph of the document containing the subject entity \textcolor{red}{Zoo Lake}. Nodes in the blue box are part of the entity graph of the document containing the object entity \textcolor{blue}{Gauteng}. Common entities are marked in orange color.}
\label{fig:graph_construction}
\end{figure*}

\subsection{Relation Classifier}

We concatenate the EGCN outputs of the nodes corresponding to the subject entity $\textbf{e}_s \in \mathbb{R}^{6(d_w+d_z)}$ and object entity $\textbf{e}_o \in \mathbb{R}^{6(d_w+d_z)}$, and pass the concatenated vector to a feed-forward network (FFN) with softmax to predict the normalized probabilities for the relation labels.
\begin{align*}
&\mathbf{r} = \mathrm{softmax}(\mathbf{W}_r (\mathbf{e}_s~||~\mathbf{e}_o) + \mathbf{b}_r)
\end{align*}
$\mathbf{W}_r \in \mathbb{R}^{ (\vert R \vert+1) \times 12(d_w+d_z)}$ is the weight matrix, $\mathbf{b}_r \in \mathbb{R}^{\vert R \vert+1}$ is the bias vector of the FFN, and $\mathbf{r}$ is the vector of normalized probabilities of relation labels.

\section{Experiments}

\subsection{Baselines}

We implement four neural baseline models for comparison with our proposed HEGCN model. Similar to our proposed model, we represent the tokens in the documents using pre-trained word embedding vectors and entity token indicator vectors. We use a document separator token when concatenating the vectors of two documents in a chain. 

(1) CNN: We apply the convolution operation on the sequence of token vectors with different kernel sizes. A max-pooling operation is applied to choose the features from the outputs of the convolution operation. This feature vector is passed to a feed-forward layer with softmax to classify the relation.

(2) BiLSTM: The token vectors of the document chain are passed to a BiLSTM layer to encode its meaning. We obtain the entity mention vectors of the subject entity and the object entity by concatenating the hidden vectors of their first and last token. We average the entity mention tokens of the corresponding entity to obtain the representation of the subject entity and the object entity. These two vectors are concatenated and passed to a feed-forward layer with softmax to find the relation between them.

(3) BiLSTM\_CNN: This is a combination of the BiLSTM and CNN model described above. The token vectors of the documents are passed to a BiLSTM layer and then we use the convolution operation with max-pooling with different convolutional kernel sizes on the hidden vectors of the BiLSTM layer. The feature vector obtained from the max-pooling operation is passed to a feed-forward layer with softmax to classify the relation.

(4) LinkPath: This model uses the explicit paths \cite{kundu2019exploiting} from the subject entity $e_s$ to the object entity $e_o$ via the common entities to find the relation. As we consider only two-hop relations, each path from $e_s$ to $e_o$ will be of the form $e_s \rightarrow c \rightarrow e_o$, where $c$ is a common entity. Since there can be multiple common entities between two documents and these common entities as well as the subject and object entities can appear multiple times in the two documents, there exist multiple paths from $e_s$ to $e_o$. Each path is formed with four entity mentions: (i) entity mentions of the subject entity and common entity in the first document. (ii) entity mentions of the common entity and object entity in the second document. We concatenate the BiLSTM hidden vectors of the start and end token of an entity mention to obtain its representation. Each path is constructed by concatenating all the four entity mentions of the path. This can be extended from two-hop to multi-hop relations by using a recurrent neural network that takes the path entity mentions as input, and outputs the hidden representation of the path. We average the vector representations of all the paths and pass it to a feed-forward layer with softmax to find the relation. 

\subsection{Parameter Settings}

We use GloVe \citep{pennington2014glove} word embeddings of dimension $d_w$ which is set to 300 in our experiments, and update the embeddings during training. We set the dimension $d_z$ to be 20 for the entity token indicator embedding vectors. The hidden vector dimension of the forward and backward LSTM is set at $320$. The dimension of BiLSTM output is $640$. We use $500$ different convolution filters with kernel width of $3$, $4$, and $5$ for feature extraction. We use one convolutional layer in both entity mention-level GCN and entity-level GCN in our final model. Dropout layers \citep{Srivastava2014DropoutAS} are used in our network with a dropout rate of $0.5$ to avoid overfitting. We train our models with a mini-batch size of $32$ and use negative log-likelihood as our objective function. We optimize the network parameters using the Adagrad optimizer \citep{duchi2011adaptive}. For evaluation, we use precision, recall, and F1 score. We do not include the {\em None} relation in the evaluation. A confidence threshold that achieves the highest F1 score on the validation dataset is used to decide if the relation of a test instance belongs to the set of relations $R$ or {\em None}. 

\subsection{Experimental Results}

We include the median of five runs of the models on the THRED dataset in Table \ref{tab:results}. We see that adding a BiLSTM in the document encoding layer improves the performance by close to 5\% in F1 score. The BiLSTM, BiLSTM\_CNN, and LinkPath models achieve similar F1 scores. When we add our proposed hierarchical entity graph convolutional layer on top of the BiLSTM layer, we get another 1.1\% F1 score improvement over the next best BiLSTM model. We perform a statistical significance test using bootstrap resampling to compare each baseline and our HEGCN model, and have ascertained that the higher F1 score achieved by our model is statistically significant ($p < 0.001$).

\begin{table}[ht]
\small
\centering
\begin{tabular}{l|ccc}
\hline
Model       & Prec. & Rec.  & F1    \\ \hline
CNN         & 0.602 & 0.655 & 0.628 \\ 
BiLSTM      & 0.682 & 0.668 & 0.675 \\ 
BiLSTM\_CNN & 0.654 & 0.696 & 0.674 \\ 
LinkPath    & 0.682 & 0.666 & 0.674 \\ \hline
HEGCN       & 0.674 & 0.699 & \textbf{0.686} \\ \hline
\end{tabular}
\caption{Performance comparison of the models on the THRED dataset. We report the median of 5 runs.}
\label{tab:results}
\end{table}

\subsection{Ablation Studies}

We include the performance of our HEGCN model with different numbers of convolutional layers in the entity mention-level GCN (EMGCN) and entity-level GCN (EGCN) in Table \ref{tab:gcn_layers_ablation}. When we increase the number of layers in either GCN, the performance of the model drops. We finally use only one convolutional layer in both EMGCN and EGCN.

\begin{table}[ht]
\small
\centering
\begin{tabular}{cc|ccc}
\hline
L1 & L2 & Prec. & Rec.  & F1    \\ \hline
1  & 1  & 0.674 & 0.699 & \textbf{0.686} \\ 
2  & 1  & 0.709 & 0.650 & 0.678 \\ 
2  & 2  & 0.682 & 0.663 & 0.673 \\ 
3  & 1  & 0.671 & 0.635 & 0.653 \\ 
3  & 2  & 0.673 & 0.667 & 0.670 \\ 
3  & 3  & 0.623 & 0.651 & 0.637 \\ \hline
\end{tabular}
\caption{The ablation study of the HEGCN model with different numbers of convolutional layers (L1 and L2) in EMGCN and EGCN.}
\label{tab:gcn_layers_ablation}
\end{table}

In Table \ref{tab:edge_ablation}, we include the ablation study of the different types of edges in EMGCN and EGCN. Removing any type of edges reduces the F1 score.

\begin{table}[ht]
\small
\centering
\begin{tabular}{l|ccc}
\hline
Model  & Prec. & Rec.  & F1    \\ \hline
HEGCN  & 0.674 & 0.699 & \textbf{0.686} \\ 
\quad -- EMG type 1  & 0.679 & 0.689 & 0.684 \\ 
\quad -- EMG type 2  & 0.698 & 0.662 & 0.680 \\ 
\quad -- EMG type 3 & 0.666 & 0.693 & 0.679 \\ 
\quad -- EG type 1 & 0.704 & 0.659 & 0.681 \\
\quad -- EG type 2 & 0.674 & 0.691 & 0.683 \\ \hline
\end{tabular}
\caption{The ablation study of the different types of edges in our HEGCN model.}
\label{tab:edge_ablation}
\end{table}

\section{Conclusion}

In this paper, we propose how the idea of distant supervision can be extended from sentence-level extraction to multi-hop extraction to cover more relations. We propose a general approach to create multi-hop relation extraction datasets. Following this approach, we create a two-hop relation extraction dataset that covers a higher number of relations from knowledge bases than other distantly supervised relation extraction datasets. We also propose a hierarchical entity graph convolutional network for this task. The two levels of GCN in our model help to capture the relation cues within documents and across documents. Our proposed model improves the F1 score by 1.1\% on our two-hop dataset, compared to a strong neural baseline, and it can be readily extended to N-hop datasets.

% \section*{Acknowledgments}

% The acknowledgments should go immediately before the references. Do not number the acknowledgments section.
% \textbf{Do not include this section when submitting your paper for review.}

\bibliographystyle{acl_natbib}
\bibliography{ranlp2021}

\begin{thebibliography}{31}
\expandafter\ifx\csname natexlab\endcsname\relax\def\natexlab#1{#1}\fi

\bibitem[{Cao et~al.(2019)Cao, Fang, and Tao}]{cao2019bag}
Yu~Cao, Meng Fang, and Dacheng Tao. 2019.
\newblock {BAG}: Bi-directional attention entity graph convolutional network
  for multi-hop reasoning question answering.
\newblock In \emph{NAACL}.

\bibitem[{De~Cao et~al.(2019)De~Cao, Aziz, and Titov}]{de2019question}
Nicola De~Cao, Wilker Aziz, and Ivan Titov. 2019.
\newblock Question answering by reasoning across documents with graph
  convolutional networks.
\newblock In \emph{NAACL}.

\bibitem[{Duchi et~al.(2011)Duchi, Hazan, and Singer}]{duchi2011adaptive}
John Duchi, Elad Hazan, and Yoram Singer. 2011.
\newblock Adaptive subgradient methods for online learning and stochastic
  optimization.
\newblock \emph{Journal of Machine Learning Research}.

\bibitem[{Gao et~al.(2019)Gao, Han, Zhu, Liu, Li, Sun, and
  Zhou}]{gao2019fewrel}
Tianyu Gao, Xu~Han, Hao Zhu, Zhiyuan Liu, Peng Li, Maosong Sun, and Jie Zhou.
  2019.
\newblock {F}ew{R}el 2.0: Towards more challenging few-shot relation
  classification.
\newblock In \emph{EMNLP and IJCNLP}.

\bibitem[{Guo et~al.(2019)Guo, Zhang, and Lu}]{guo2019aggcn}
Zhijiang Guo, Yan Zhang, and Wei Lu. 2019.
\newblock Attention guided graph convolutional networks for relation
  extraction.
\newblock In \emph{ACL}.

\bibitem[{Hochreiter and Schmidhuber(1997)}]{hochreiter1997long}
Sepp Hochreiter and J{\"u}rgen Schmidhuber. 1997.
\newblock Long short-term memory.
\newblock \emph{Neural Computation}.

\bibitem[{Hoffmann et~al.(2011)Hoffmann, Zhang, Ling, Zettlemoyer, and
  Weld}]{hoffmann2011knowledge}
Raphael Hoffmann, Congle Zhang, Xiao Ling, Luke Zettlemoyer, and Daniel~S Weld.
  2011.
\newblock Knowledge-based weak supervision for information extraction of
  overlapping relations.
\newblock In \emph{ACL}.

\bibitem[{Jat et~al.(2017)Jat, Khandelwal, and Talukdar}]{jat2018attention}
Sharmistha Jat, Siddhesh Khandelwal, and Partha Talukdar. 2017.
\newblock Improving distantly supervised relation extraction using word and
  entity based attention.
\newblock In \emph{AKBC}.

\bibitem[{Kipf and Welling(2017)}]{Kipf2017SemiSupervisedCW}
Thomas Kipf and Max Welling. 2017.
\newblock Semi-supervised classification with graph convolutional networks.
\newblock In \emph{ICLR}.

\bibitem[{Kundu et~al.(2019)Kundu, Khot, Sabharwal, and
  Clark}]{kundu2019exploiting}
Souvik Kundu, Tushar Khot, Ashish Sabharwal, and Peter Clark. 2019.
\newblock Exploiting explicit paths for multi-hop reading comprehension.
\newblock In \emph{ACL}.

\bibitem[{Lin et~al.(2016)Lin, Shen, Liu, Luan, and Sun}]{lin2016neural}
Yankai Lin, Shiqi Shen, Zhiyuan Liu, Huanbo Luan, and Maosong Sun. 2016.
\newblock Neural relation extraction with selective attention over instances.
\newblock In \emph{ACL}.

\bibitem[{Mintz et~al.(2009)Mintz, Bills, Snow, and
  Jurafsky}]{mintz2009distant}
Mike Mintz, Steven Bills, Rion Snow, and Dan Jurafsky. 2009.
\newblock Distant supervision for relation extraction without labeled data.
\newblock In \emph{ACL and IJCNLP}.

\bibitem[{Nayak et~al.(2021)Nayak, Majumder, Goyal, and
  Poria}]{Nayak2021DeepNA}
Tapas Nayak, Navonil Majumder, Pawan Goyal, and Soujanya Poria. 2021.
\newblock Deep neural approaches to relation triplets extraction: A
  comprehensive survey.
\newblock \emph{Cognitive Computing}.

\bibitem[{Nayak and Ng(2019)}]{nayak2019effective}
Tapas Nayak and Hwee~Tou Ng. 2019.
\newblock Effective attention modeling for neural relation extraction.
\newblock In \emph{CoNLL}.

\bibitem[{Nayak and Ng(2020)}]{nayak2019ptrnetdecoding}
Tapas Nayak and Hwee~Tou Ng. 2020.
\newblock Effective modeling of encoder-decoder architecture for joint entity
  and relation extraction.
\newblock In \emph{AAAI}.

\bibitem[{Pennington et~al.(2014)Pennington, Socher, and
  Manning}]{pennington2014glove}
Jeffrey Pennington, Richard Socher, and Christopher Manning. 2014.
\newblock {GloVe}: Global vectors for word representation.
\newblock In \emph{EMNLP}.

\bibitem[{Riedel et~al.(2010)Riedel, Yao, and McCallum}]{riedel2010modeling}
Sebastian Riedel, Limin Yao, and Andrew McCallum. 2010.
\newblock Modeling relations and their mentions without labeled text.
\newblock In \emph{ECML and KDD}.

\bibitem[{Shen and Huang(2016)}]{huang2016attention}
Yatian Shen and Xuanjing Huang. 2016.
\newblock Attention-based convolutional neural network for semantic relation
  extraction.
\newblock In \emph{COLING}.

\bibitem[{Song et~al.(2018)Song, Wang, Yu, Zhang, Florian, and
  Gildea}]{song2018exploring}
Linfeng Song, Zhiguo Wang, Mo~Yu, Yue Zhang, Radu Florian, and Daniel Gildea.
  2018.
\newblock Exploring graph-structured passage representation for multi-hop
  reading comprehension with graph neural networks.
\newblock \emph{CoRR}.

\bibitem[{Srivastava et~al.(2014)Srivastava, Hinton, Krizhevsky, Sutskever, and
  Salakhutdinov}]{Srivastava2014DropoutAS}
Nitish Srivastava, Geoffrey~E. Hinton, Alex Krizhevsky, Ilya Sutskever, and
  Ruslan Salakhutdinov. 2014.
\newblock Dropout: a simple way to prevent neural networks from overfitting.
\newblock \emph{JMLR}.

\bibitem[{{Takanobu} et~al.(2019){Takanobu}, {Zhang}, {Liu}, and
  {Huang}}]{takanobu2019hrlre}
Ryuichi {Takanobu}, Tianyang {Zhang}, Jiexi {Liu}, and Minlie {Huang}. 2019.
\newblock A hierarchical framework for relation extraction with reinforcement
  learning.
\newblock In \emph{AAAI}.

\bibitem[{Vashishth et~al.(2018)Vashishth, Joshi, Prayaga, Bhattacharyya, and
  Talukdar}]{vashishth2018reside}
Shikhar Vashishth, Rishabh Joshi, Sai~Suman Prayaga, Chiranjib Bhattacharyya,
  and Partha Talukdar. 2018.
\newblock {RESIDE}: Improving distantly-supervised neural relation extraction
  using side information.
\newblock In \emph{EMNLP}.

\bibitem[{Vrande{\v{c}}i{\'c} and Kr{\"o}tzsch(2014)}]{vrandevcic2014wikidata}
Denny Vrande{\v{c}}i{\'c} and Markus Kr{\"o}tzsch. 2014.
\newblock Wikidata: a free collaborative knowledgebase.
\newblock \emph{Communications of Association for Computing Machinery}.

\bibitem[{Welbl et~al.(2018)Welbl, Stenetorp, and
  Riedel}]{welbl2018constructing}
Johannes Welbl, Pontus Stenetorp, and Sebastian Riedel. 2018.
\newblock Constructing datasets for multi-hop reading comprehension across
  documents.
\newblock \emph{TACL}.

\bibitem[{Yao et~al.(2019)Yao, Ye, Li, Han, Lin, Liu, Liu, Huang, Zhou, and
  Sun}]{yao2019DocRED}
Yuan Yao, Deming Ye, Peng Li, Xu~Han, Yankai Lin, Zhenghao Liu, Zhiyuan Liu,
  Lixin Huang, Jie Zhou, and Maosong Sun. 2019.
\newblock {DocRED}: A large-scale document-level relation extraction dataset.
\newblock In \emph{ACL}.

\bibitem[{Ye and Ling(2019)}]{ye2019distant}
Zhi-Xiu Ye and Zhen-Hua Ling. 2019.
\newblock Distant supervision relation extraction with intra-bag and inter-bag
  attentions.
\newblock In \emph{NAACL}.

\bibitem[{Zeng et~al.(2015)Zeng, Liu, Chen, and Zhao}]{zeng2015distant}
Daojian Zeng, Kang Liu, Yubo Chen, and Jun Zhao. 2015.
\newblock Distant supervision for relation extraction via piecewise
  convolutional neural networks.
\newblock In \emph{EMNLP}.

\bibitem[{Zeng et~al.(2014)Zeng, Liu, Lai, Zhou, and Zhao}]{zeng2014relation}
Daojian Zeng, Kang Liu, Siwei Lai, Guangyou Zhou, and Jun Zhao. 2014.
\newblock Relation classification via convolutional deep neural network.
\newblock In \emph{COLING}.

\bibitem[{Zeng et~al.(2020)Zeng, Xu, Chang, and Li}]{Zeng2020DoubleGB}
Shuang Zeng, Runxin Xu, Baobao Chang, and Lei Li. 2020.
\newblock Double graph based reasoning for document-level relation extraction.
\newblock In \emph{EMNLP}.

\bibitem[{Zhang et~al.(2018)Zhang, Qi, and Manning}]{zhang2018graph}
Yuhao Zhang, Peng Qi, and Christopher~D. Manning. 2018.
\newblock Graph convolution over pruned dependency trees improves relation
  extraction.
\newblock In \emph{EMNLP}.

\bibitem[{Zhang et~al.(2017)Zhang, Zhong, Chen, Angeli, and
  Manning}]{zhang2017position}
Yuhao Zhang, Victor Zhong, Danqi Chen, Gabor Angeli, and Christopher~D.
  Manning. 2017.
\newblock Position-aware attention and supervised data improve slot filling.
\newblock In \emph{EMNLP}.

\end{thebibliography}

%\appendix

\end{document}